\documentclass{article}

\PassOptionsToPackage{numbers, compress}{natbib}
\usepackage[final]{nips}

\usepackage[utf8]{inputenc} 
\usepackage[T1]{fontenc}    
\usepackage{hyperref}       
\usepackage{url}            
\usepackage{booktabs}       
\usepackage{amsfonts}       
\usepackage{nicefrac}       
\usepackage{microtype}      

\usepackage{epsfig}
\usepackage{graphicx}
\usepackage{amsmath}
\usepackage{amssymb}
\usepackage{algorithm,algorithmic}
\usepackage{subfigure} 
\usepackage{multirow}
\usepackage{bm}
\usepackage{hyphenat}

\title{Unified Backpropagation \\ for Multi-Objective Deep Learning}

\author{
  Arash shahriari \\
  Research School of Engineering \\
  Australian National University \\
  Canberra, Australia \\
  \texttt{arash.shahriari@anu.edu.au} \\
}

\begin{document}

\maketitle

\begin{abstract}
\nohyphens{A common practice in most of deep convolutional neural architectures is to employ fully-connected layers followed by Softmax activation to minimize cross-entropy loss for the sake of classification. Recent studies show that substitution or addition of the Softmax objective to the cost functions of support vector machines or linear discriminant analysis is highly beneficial to improve the classification performance in hybrid neural networks. We propose a novel paradigm to link the optimization of several hybrid objectives through unified backpropagation. This highly alleviates the burden of extensive boosting for independent objective functions or complex formulation of multiobjective gradients. Hybrid loss functions are linked by basic probability assignment from evidence theory. We conduct our experiments for a variety of scenarios and standard datasets to evaluate the advantage of our proposed unification approach to deliver consistent improvements into the classification performance of deep convolutional neural networks.}
\end{abstract}

\section{Introduction}

A common practice in most deep convolutional neural networks, is to employ fully-connected layers, followed by a Softmax activation to minimize cross-entropy loss. Recent studies has shown that, substitution of the Softmax objective with SVM or LDA cost functions, is highly effective to improve the classification performance of deep neural networks. This paper proposes a novel paradigm to link the optimization of several objectives through a unified backpropagation scheme. This alleviates the burden of extensive boosting for each independent objective functions and avoids complex formulation of multi-objective gradients. Here, several loss functions are linked through BPA at the time of backpropagation. 

Deep learning has been proven to be extremely successful for several applications. The combination of machine learning methods, with deep neural networks, achieves better performances. Deep versions of CCA~(\cite{andrew2013deep}), FA~(\cite{clevert2015rectified}), PCA~(\cite{chan2015pcanet}), SVM~(\cite{vinyals2012learning}), and finally, LDA~(\cite{stuhlsatz2012feature}), have been introduced in the literature. There are two schools of thought about how to alternate the Softmax layer, so as, to achieve better performance. 

The first strategy trains a deep architecture to produce high-order features and give them to a classifier~(\cite{coates2010analysis}). For example, replacing the Softmax with SVM as the top layer and minimizing of a standard hinge loss, produces better results in some deep architectures~(\cite{tang2013deep}). Another successful practice is LDA which maximizes an objective, derived from a general eigenvalue problem~(\cite{dorfer2015deep}). The drawback is that, the features in the bottom layers are not further fine-tuned with the new objective. 

\newpage

The second strategy trains a combination of objectives, by error backpropagation through the gradients of their loss functions. It is possible to optimize these objectives with either boosting methods or multi-objective evolutionary algorithms~(\cite{gong2015multiobjective}), but the former needs extensive training of different networks, and the latter requires very complex formulations for the gradients.

\section{Method}
\label{sec:object:multiobjective:method}

A novel unified backpropagation scheme is proposed for deep multi-objective learning, based on BPA of evidence theory, applying to a network that includes different loss functions, such as, Softmax, SVM, and LDA. In contrast to the boosting method, which trains each loss function independently to make an ensemble of models, the proposed backpropagation approach, unifies the gradients of all objective functions. 

The advantages of unified backpropagation can be outlined as follows. First, this scheme mutually optimizes all the objective functions together. In this way, the contribution of each objective function to the overall classification performance, is managed by sharing the basic probability masses, among the gradients. Second, this unification is less computationally expensive than ensemble learning. Third, it prevents more complexity on the formulation of gradients, for each of the combined loss functions. The experiments for a variety of scenarios and standard datasets, confirm the advantage of the proposed approach, which delivers consistent improvements to the performance of deep convolutional neural networks.

\begin{figure}[!t]
\begin{center}
\includegraphics[width=1.0\textwidth]{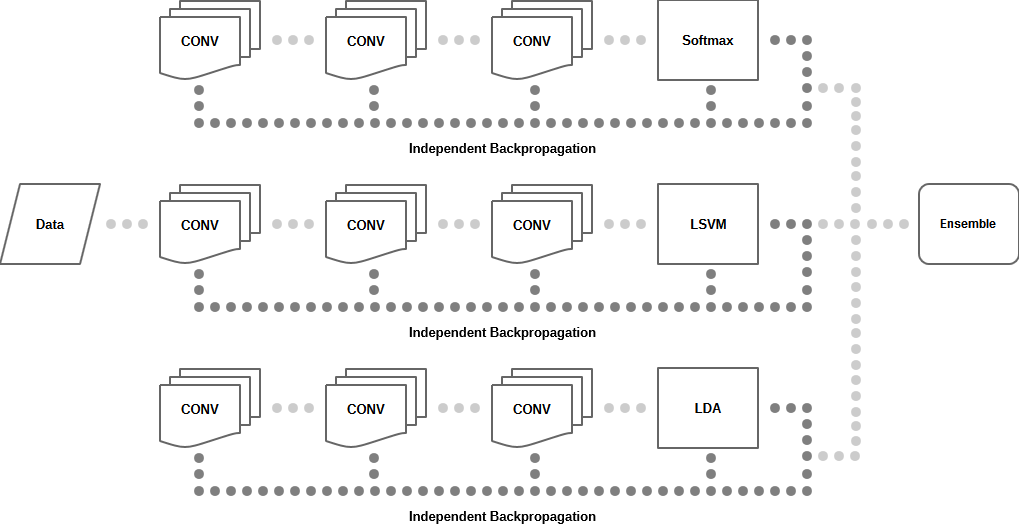}
\end{center}
\caption[Multi-objective learning of deep convolutional neural networks.]{Multi-objective learning of deep convolutional neural networks. This follows the common practice of backpropagation in identical convolutional architectures and minimizing some specific loss functions, via stochastic gradient descent. The trained deep classifiers are then boosted by a proper ensemble method. The bottleneck results from the need for multiple instances of training for each of the objective functions. The backpropagation procedures are thoroughly independent, because each of the loss functions minimize by a specific gradient formulation, with zero awareness of other ongoing training processes.}
\label{hybrid}
\end{figure}

\section{Formulation}
\label{sec:object:multiobjective:formulation}

A deep convolutional architecture can boost the typical Softmax layer with other classifiers, using a multi-objective optimization regime. The two widely-used classifiers are SVM and LDA, employed as the top layer of the deep neural networks.

\newpage

Employing SVM, a deep convolutional network optimizes the primal problem of SVM and backpropagate the gradients of the top layer SVM, to learn the bottom layers. This is in contrast with fine-tuning, where low-order features are usually trained by Softmax, and top layers tuned by SVM. It has been shown that the performance on standard benchmarks, is much better than the networks with Softmax layer at top. The optimization is performed using stochastic gradient descent method~(\cite{chan2015pcanet}). 

For LDA as the top layer, the deep architecture is almost the same. The objective of a deep neural network is reformulated to learn linearly separable features by backpropagation, because LDA allows to define optimal linear decision boundaries in the latent layers. This finds linear combinations of low-level features, to maximize the scattering between classes of data, whilst minimizing the discrepancy within individual classes. The top layer LDA tries to produce high separation between deep features, rather than, minimizing the norm of prediction error~(\cite{dorfer2015deep}). 

Classification of some multiclass datasets, is challenging due to non-uniform distribution of data~(\cite{kocco2013multi}). The accuracy of classification does not seem to be a suitable objective to optimize, because there may be high accuracy with strong biases towards some classes~(\cite{fawcett2006introduction}). Although there are many algorithms to deal with imbalanced data for binary classification~(\cite{he2009learning}), the multiclass problem has been usually addressed by generalization of the binary solutions, with a one-versus-all strategy~(\cite{abe2004iterative}). For some leaning tasks, optimization of some relevant measures within the imbalance data distribution, provides alternative measures to the accuracy~(\cite{wang2012multiclass}). 

The emergence of new cost-sensitive methods for dealing with imbalanced multiclass data~(\cite{elkan2001foundations}), has enabled the embedding of misclassification costs, into a cost matrix. They usually measure the error, based on misclassification costs of individual classes in the confusion matrix. This matrix is the most informative contingency table in multiclass learning problems, because it gives the success rate of a classifier for a special class, and the failure rate on distinguishing that class from other classes. The confusion matrix has proven to be a great regularizer; smoothing the accuracy among classes~(\cite{ralaivola2012confusion}).

The determination of a probabilistic distribution from the confusion matrix is highly effective at producing a probability assignment, which contributes to imbalance distribution problems. The probability assignments can be constructed from recognition, substitution and rejection rates~(\cite{xu1992methods}), or both precision and recall rates~(\cite{deng2016improved}). The key point is to harvest the maximum possible prior knowledge, provided by the confusion matrix to overcome the imbalance classification challenge.

\begin{figure}[!t]
\begin{center}
\includegraphics[width=1.0\textwidth]{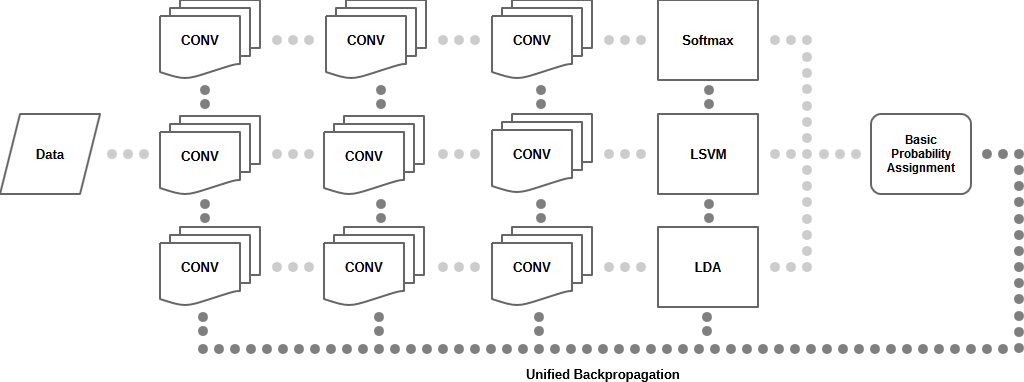}
\end{center}
\caption[Unified backpropagation for multi-objective learning.]{Unified backpropagation for multi-objective learning. In contrast with the ensemble practice, the backpropagation is unified through BPA. Although each deep convolutional network holds its specific gradient formulation, sharing of probability masses, makes a common sense around the statistics of gradient descent in the whole network. In other words, boosting procedure not only includes the outcome of several objective functions, but also enhances their mutual learning processes, by unifying their backpropagation.}
\label{unified}
\end{figure}

\newpage

\subsection{Basic Probability Assignment}
\label{sec:assignment}

\begin{algorithm}[!t]
\caption{Basic Probability Assignment}
\label{alg:bpa}
\begin{algorithmic}
\STATE
\STATE {\bfseries Input:} training set $\mathcal{X}$, classifier $\phi$
\STATE {\bfseries Output:} basic probability assignment $\bm{\Theta}_{\phi}$
\STATE
\STATE 1: Compute recall and precision~(Equation~\ref{eq:object:14})
\STATE 2: Calculate recall and precision assignments (Equation~\ref{eq:object:15})
\STATE 3: Apply Dempster-Schafer rule of combination (Equation~\ref{eq:object:16})
\STATE
\end{algorithmic}
\end{algorithm}

A confusion matrix is generally represented as class-based predictions against actual labels, in the form of a square matrix. Inspired by Dempster-Schafer theory~(\cite{sentz2002combination}), construction of BPA gives a vector, which is independent of the number of samples in each classes and sums up to one, for each labels. BPA provides the ability to reflect the different contributions of a classifier, or combine the outcomes of multiple weak classifiers. A raw, two-dimensional confusion matrix, indexed by the predicted classes and actual labels, provides some common measures of classification performance. Some general measures are accuracy (the proportion of the total number of predictions that are correct), precision (a measure of the accuracy, provided that a specific class has been predicted), recall (a measure of the ability of a prediction model to select instances of a certain class from a dataset), and F-score (the harmonic mean of precision and recall)~(\cite{sammut2011encyclopedia}).

Suppose that a set of training samples $\mathcal{X}=\{X_1,\dots,X_{|\mathcal{X}|}\}$ from $\mathcal{C}=\{C_1,\dots,C_{|\mathcal{C}|}\}$ different classes, are assigned to a label set $\mathcal{L}=\{L_1,\dots,L_{|\mathcal{L}|}\}$, using a classifier $\phi$. Each element $n_{ij}$ of the confusion matrix $\bm{\Upsilon}_{\phi}$ is considered as the number of samples belonging to class $C_i$, which assigned to label $L_j$. The recall ($r_{ij}$) and precision ($s_{ij}$) ratios for all $i\in\{1\dots{|\mathcal{C}|}\}$ and $j\in\{1\dots{|\mathcal{L}|}\}$, can be defined as follows~(\cite{deng2016improved}),

\begin{eqnarray}
r_{ij} &=& \frac{n_{ij}}{\sum_{i=1}^{|\mathcal{C}|}{n_{ij}}}\nonumber\\
s_{ij} &=& \frac{n_{ij}}{\sum_{j=1}^{|\mathcal{L}|}{n_{ij}}}
\label{eq:object:14}
\end{eqnarray}

It can be seen that, the recall ratio is summed over the predicted classes (rows), whilst the precision ratio is accumulated by the actual labels (columns) of the confusion matrix $\bm{\Upsilon}_{\phi}$. The probability elements of recall ($R_{i}$) and precision ($S_{i}$) for each individual class $C_i$ are, 

\begin{eqnarray}
R_{i}&=&\frac{r_{ii}}{\sum_{j=1}^{|\mathcal{C}|}{r_{jj}}}\nonumber\\
S_{i}&=&\frac{s_{ii}}{\sum_{j=1}^{|\mathcal{L}|}{s_{jj}}} 
\label{eq:object:15}
\end{eqnarray}

These elements are synthesized to form the final probability assignments by Dempster-Schafer rule of combination~(\cite{sentz2002combination}), representing the recognition ability of classifier $\phi$ to each of the classes of set $\mathcal{C}$ as,

\begin{equation}
M_i = R_i\oplus{S_i} = \frac{R_i{S_i}}{1-\sum_{i=1}^{|\mathcal{C}|}R_i{S_i}}
\label{eq:object:16}
\end{equation}

\noindent where the operator $\oplus$ is an orthogonal sum. The overall contribution of the classifier $\phi$ can be presented as a probability assignment vector,

\begin{equation}
\bm{\Theta}_{\phi}=\{M_i\;|\;\forall\;i\in[1,|\mathcal{C}|]\}
\label{eq:object:17}
\end{equation}

It is worth mentioning that $\bm{\Theta}_{\phi}$ should be computed by the training set, because it is assumed that, there is no actual label set $\mathcal{L}$ at the test time.

\newpage

\subsection{Unified Backpropagation}
\label{sec:object:multiobjective:transfer}

Suppose that $\Phi=\{\phi_1,\dots,\phi_{|\Phi|}\}$ is a set of objectives in a multi-objective learning regime, presented in Figure~\ref{unified}. To apply the unified backpropagation, Algorithm~\ref{alg:bpa} is deployed for each of the loss functions, to come up with a set of normalized probability assignments as,

\begin{equation}
\bm{\Gamma}_{\Phi}=\{\bm{\Gamma}_{\phi_k}=\lVert{\bm\Theta_{\phi_k}}\rVert_2\;|\;\forall\;k\in[1,|\Phi|]\}
\label{eq:object:23}
\end{equation}

\noindent where $\bm\Theta_{\phi_k}$ follows the same definition as Equation~\ref{eq:object:17}.

In each layer $l$ of the $k$th objective, feedforward propagation is calculated as follows,

\begin{equation}
a_k^{(l)} = \sigma(w_k^{(l)}{a_k^{(l-1)}}+b_k^{(l)})
\label{eq:object:24}
\end{equation}

\noindent where $w_k$ and $b_k$ are weights and biases, $a_k$ is an activation and $\sigma$ is a rectification function. Considering $\phi_k$ as the loss function, the output error holds,

\begin{equation}
\delta_k^{(l)} = \bigtriangledown_{a_k}\phi_k\odot\sigma^{'}(w_k^{(l)}{a_k^{(l-1)}}+b_k^{(l)})
\label{eq:object:25}
\end{equation}

The backpropagation error can be stated as,

\begin{equation}
\delta_k^{(l)} = ((w_k^{(l+1)})^T)\delta_k^{(l+1)}\odot\sigma^{'}(w_k^{(l)}{a_k^{(l-1)}}+b_k^{(l)})
\label{eq:object:26}
\end{equation}

For the sake of gradient descent, the weights and biases are updated via,

\begin{eqnarray}
w_k^{(1)}&\longrightarrow&w_k^{(1)}-\eta\;\Gamma_{\phi_k}\sum\delta_k^{(l+1)}(a_k^{(l-1)})^T\nonumber\\
b_k^{(1)}&\longrightarrow&b_k^{(1)}-\eta\;\Gamma_{\phi_k}\sum\delta_k^{(l+1)}
\label{eq:object:27}
\end{eqnarray}

\begin{algorithm}[tb]
   \caption{Unified Backpropagation}
   \label{alg:object:multiobjective:unified}
\begin{algorithmic}
   	\STATE
	\STATE {\bfseries Input:} set of objective functions $\Phi$ 
	\STATE {\bfseries Output:} multi-objective network
	\STATE
	\STATE 1: Compute $\bm{\Gamma}_{\Phi}$~(Equation~\ref{eq:object:23}) by using Algorithm~\ref{alg:bpa}
	\STATE
	\FOR{$k=1$ {\bfseries to} $|\Phi|$}
	\FORALL{layers}
	\STATE 2: Compute feadforward activations~(Equation~\ref{eq:object:24})
	\STATE 3: Calculate backpropagation errors~(Equation~\ref{eq:object:26})
	\STATE 4: Update weights and biases~(Equation~\ref{eq:object:27})
	\ENDFOR
	\ENDFOR
	\STATE
\end{algorithmic}
\end{algorithm}

For the unified backpropagation, larger $\Gamma_{\phi_k}$ of the $k$th objective will generate bigger update rates for weights and biases, than only employing a fix $\eta$. This helps to update only loss functions, which largely affect the overall classification performance, and properly connect the objectives in the backpropagation process. This also implies that, in spite of forward-backward propagation, the overall contribution of each objective function is taken into account. Algorithm~\ref{alg:object:multiobjective:unified} wraps up the proposed unification strategy.

\newpage

\subsection{Objective Functions}
\label{sec:object:multiobjective:objectives}

Following the successful practices in the literature, three types of widely-used loss functions \textit{i.e.} Softmax, SVM~(\cite{tang2013deep}), and LDA~(\cite{dorfer2015deep}) are further investigated. Suppose that $\mathcal{C}=\{C_1,\dots,C_{|\mathcal{C}|}\}$ is a set of $|\mathcal{C}|$ different classes in the dataset at hand and the discrete probability distribution $p_i$ denotes, to what extent, each sample of set $\mathcal{X}=\{X_1,\dots,X_{|\mathcal{X}|}\}$ belongs to class $C_i$. Assuming $a_k=\{a_{k_1},\dots,a_{k_{|\mathcal{C}|}}\}$ as the output of the last fully-connected layer for the $k$th loss function (Equation~\ref{eq:object:24}), the closed form formulation of gradients for the above objective functions, can be worked out as follows.   

\subsubsection{Softmax}
\label{sec:Softmax}

For a conventional Softmax activation, $p_i$ can be defined as follows,

\begin{equation}
p_i = \frac{e^{a_{k_i}}}{\sum_{j=1}^{|\mathcal{C}|}e^{a_{k_j}}}
\label{eq:object:28}
\end{equation}

\noindent such that $\sum{p_i}=1$ and the predicted class is yielded by,

\begin{equation}
y_i = \arg\max\;p_i=\arg\max\;a_i
\label{eq:object:29}
\end{equation}

The Softmax loss function forms as,

\begin{equation}
\phi_{k_{Softmax}} = -\sum_{j=1}^{|\mathcal{C}|}y_j\;log(p_j)
\label{eq:object:30}
\end{equation}

\noindent where its gradient with respect to $a_{k_i}$ holds,

\begin{equation}
\bigtriangledown_{a_{k_i}}\phi_{k_{Softmax}}=p_i-y_i
\label{eq:object:31}
\end{equation}

\subsubsection{Support Vector Machine}
\label{sec:object:multiobjective:svm}

The squared hinge loss for $L_2$-norm binary SVM ($t_i\in\{-1,+1\}$) is defined as,

\begin{equation}
\phi_{k_{SVM}} = \frac{1}{2}\bm{w}^T\bm{w}+\lambda\sum_{i=1}^{|\mathcal{C}|}max(0, 1-\bm{w}^T{a_{k_i}}{t_i})^2
\label{eq:object:32}
\end{equation}

\noindent that its gradient can be derived as follows,

\begin{equation}
\bigtriangledown_{a_{k_i}}\phi_{k_{SVM}}=-2\lambda{t_i}\bm{w}\big(max(0, 1-\bm{w}^T{a_{k_i}}{t_i})\big)
\label{eq:object:33}
\end{equation}

The multiclass scenario is the extension of the binary objective, using one-vs-rest approach. Minimizing Equation~\ref{eq:object:33} for $\bm{w}$, gives the predicted class as,

\begin{equation}
y_i=\arg\max\;{w_i}{a_{k_i}}\;\;\;\forall\;{w_i}\in{\bm{w}}
\label{eq:object:34}
\end{equation}

\newpage

\subsubsection{Linear Discriminant Analysis}
\label{sec:object:multiobjective:lda}

The focus is on maximizing the $|\mathcal{C}|-1$ smallest eigenvalues of the generalized LDA eigenvalue problem, 

\begin{equation}
{\bm{S}_b}{\bm{e}}={\bm{v}{\bm{S}_w}{\bm{e}}}
\label{eq:object:35}
\end{equation}

\noindent such that, $\bm{S}_b$ is between-class scattering, $\bm{S}_w$ is within-class scattering, $\bm{e}$ corresponds to eigenvectors and $\bm{v}$ represents the eigenvalues. This leads to a maximization of the discriminant power of any deep architecture. Hence, the objective can be stated as,

\begin{equation}
\phi_{k_{LDA}}=\frac{1}{|\mathcal{C}|-1}\sum_{i=1}^{|\mathcal{C}|-1}{v_i}\;\;\;\forall\;{v_i}\in{\bm{v}}
\label{eq:object:36}
\end{equation}

\noindent which holds the following gradient ($\forall{e_i}\in{\bm{e}}$),

\begin{equation}
\bigtriangledown_{a_{k_i}}\phi_{k_{LDA}}=\frac{1}{|\mathcal{C}|-1}\sum_{i=1}^{|\mathcal{C}|-1}{e_i}^T\bigg(\frac{\delta{\bm{S}_b}}{\delta{a_{k_i}}}-\frac{\delta{\bm{S}_w}}{\delta{a_{k_i}}}\bigg){e_i}
\label{eq:object:37}
\end{equation}

\section{Experiments}
\label{sec:object:multiobjective:experiments}

The experiments are conducted for two different scenarios, using MNIST~(\cite{lecun1998gradient}), CIFAR~(\cite{krizhevsky2009learning}), and SVHN~(\cite{netzer2011reading}) datasets. In the first scenario, the unified backpropagation is applied to single-objective learning and its performance is compared to the baseline of Softmax, SVM and LDA. For the second scenario, multi-objective learning are considered and multiple loss functions are combined via the unification backpropagation paradigm. We report the results on standard architectures, implemented in deep learning library of the Oxford Visual Geometry Group~(\cite{vedaldi08vlfeat}).

\subsection{Single-Objective Learning}
\label{sec:single}

In this scenario, the advantage of the unified backpropagation (Unified) is validate for each of the individual objective functions under examination. We provide the outcomes for the Softmax, SVM, and LDA as common loss functions among deep convolutional networks.

Tables~\ref{tab:single-loss-Softmax},~\ref{tab:single-loss-svm}, and~\ref{tab:single-loss-lda} show the outcomes of this scenario. It can be seen that, the unified backpropagation is able to consistently improve the classification performance on deep convolutional networks, and provides smaller training-test errors. According to the Table~\ref{tab:single-loss-Softmax}, the unified backpropagation outperforms all the baselines. The greatest improvement belongs to CIFAR-10, which reduces the test error from 22.72\% to 18.77\%. The smallest improvement on the training error, comes from the same datasets. It seems that, in spite of larger training error, better generalization leads to considerable enhancement in the overall performance.

In Table~\ref{tab:single-loss-svm}, the best improvement in test error goes to CIFAR-100, which decreases from 49.11\% to 39.76\%. Although the number of classes are higher for other datasets, the unified backpropagation successfully avoids biases for SVM and hence, the overall performance is considerably improved. Looking at the Table~\ref{tab:single-loss-lda}, the best result is recorded for MNIST. The unification paradigm outperforms in both training and test errors, on all the experimental datasets. This is the result of distinction, imposed by LDA. Since LDA pushes the separation among classes, rather than the likelihood of predictions and labels (Softmax,SVM), the training-test errors reduce accordingly in all the experiments. This confirms that the proposed scheme is successful at providing better learning practices, compared to the typical methods.

\begin{figure}[p]
\begin{center}
\includegraphics[width=0.6\textwidth]{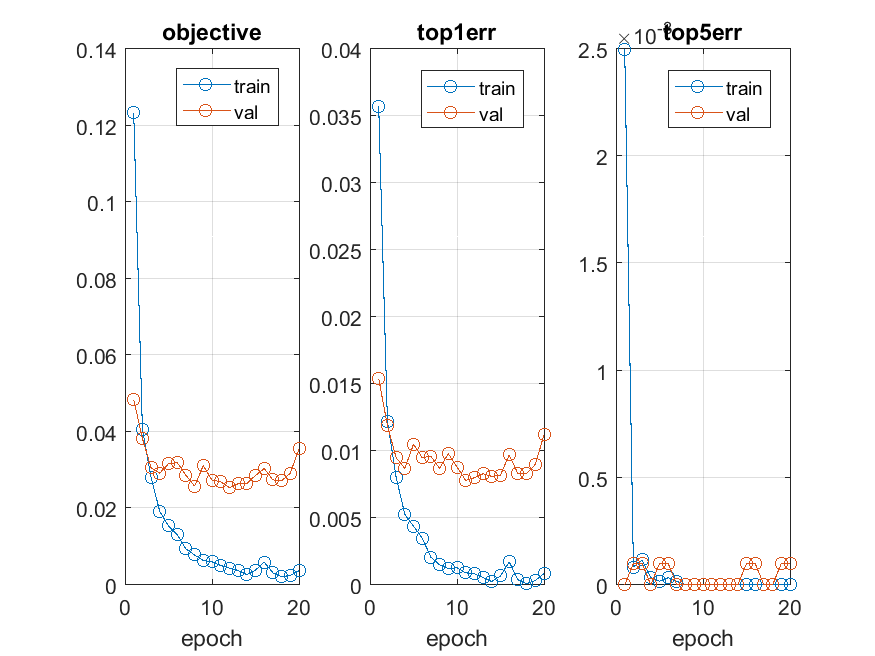}
\end{center}
\begin{center}
\includegraphics[width=0.6\textwidth]{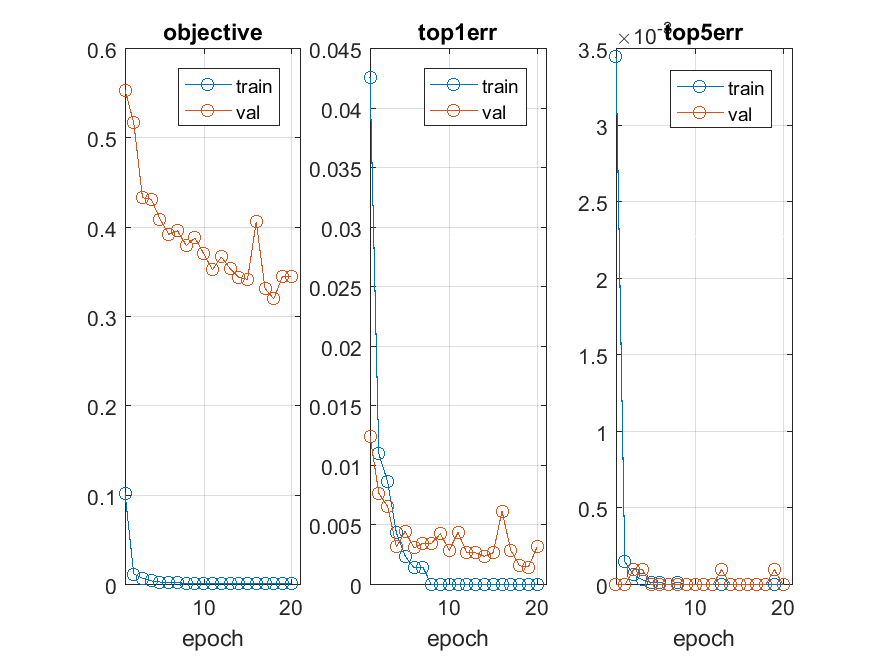}
\end{center}
\begin{center}
\includegraphics[width=0.6\textwidth]{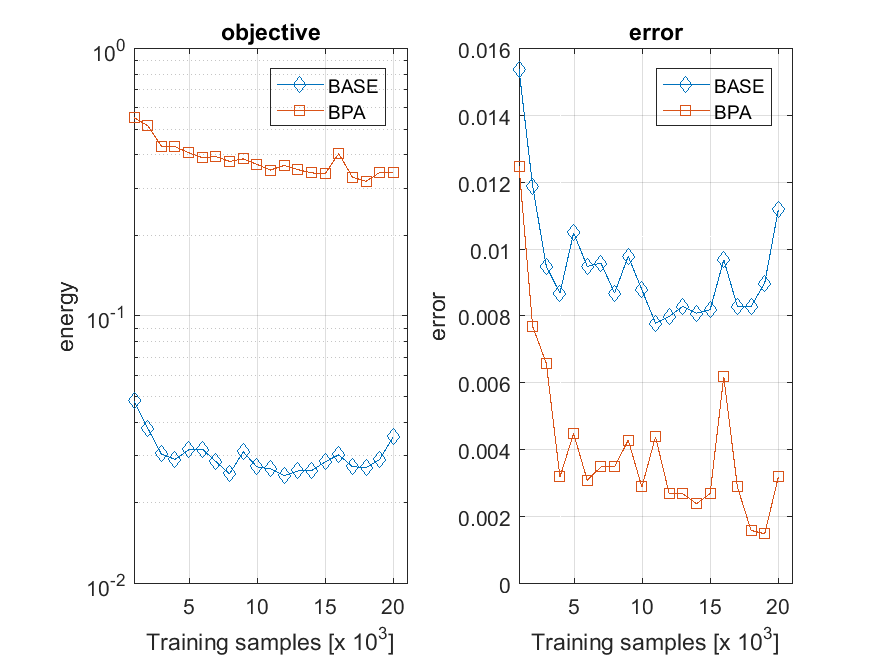}
\end{center}
\caption[Unified backpropagation for MNIST.]{Unified backpropagation for MNIST dataset. For Softmax baseline (chart 1), the training-validation losses and top-1 errors, follow each other accordingly, but the gap between them is fairly wide. By the unification approach (chart2), the validation gives higher energy than the training, but the errors are considerably smaller than the baseline. This also provides better generalization, due to the closer gap between training-validation errors. In spite of the higher level of energy (chart 3), the unified backpropagation consistently improves the precision of Softmax.}
\label{mnist}
\end{figure}

\begin{figure}[p]
\begin{center}
\includegraphics[width=0.6\textwidth]{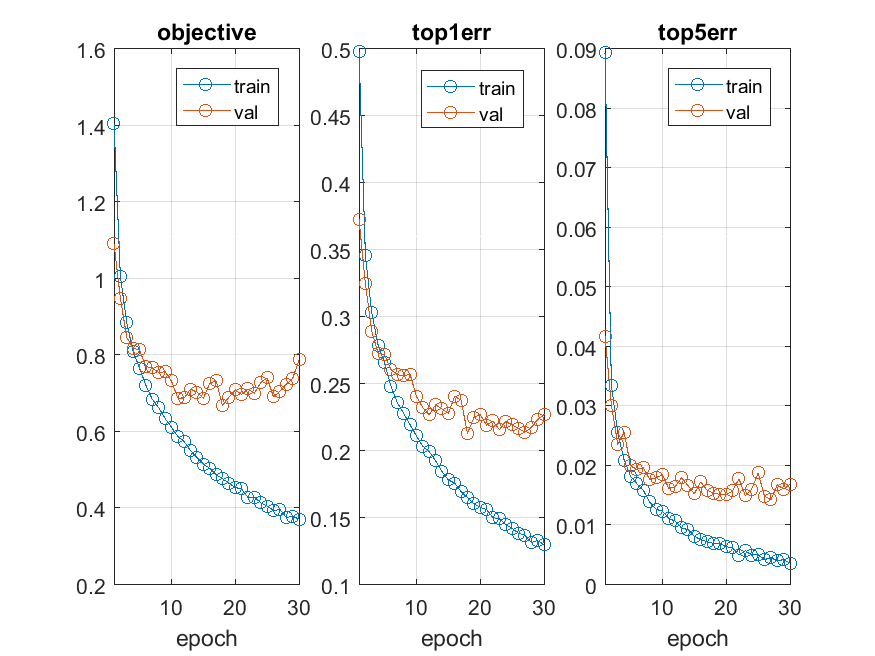}
\end{center}
\begin{center}
\includegraphics[width=0.6\textwidth]{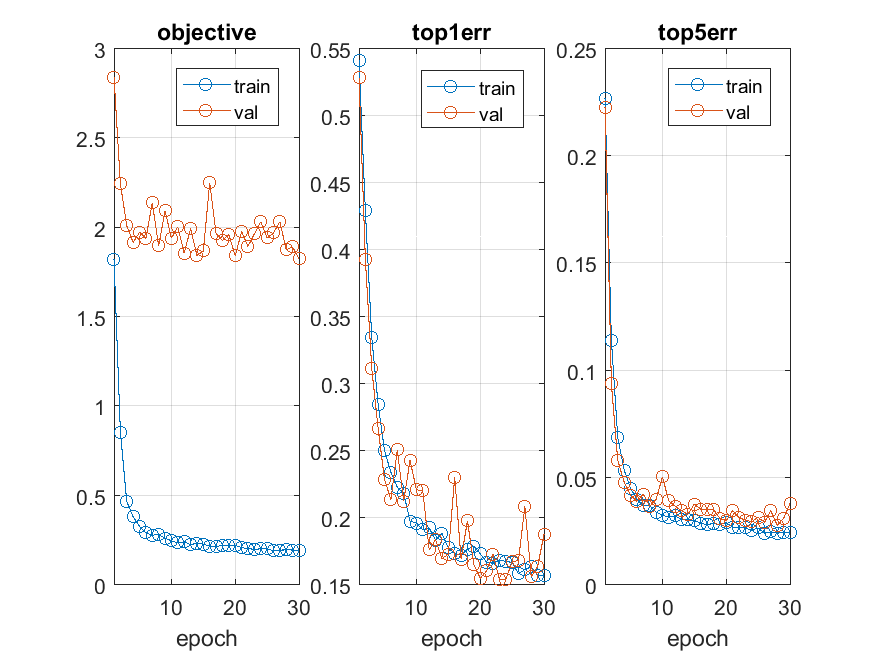}
\end{center}
\begin{center}
\includegraphics[width=0.6\textwidth]{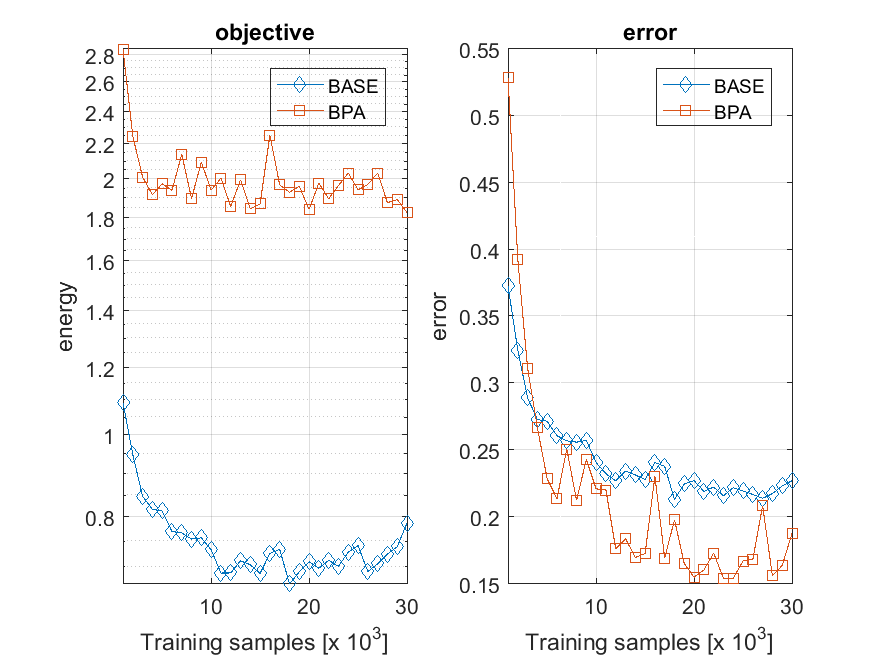}
\end{center}
\caption[Unified backpropagation for CIFAR-10.]{Unified backpropagation for CIFAR-10 dataset. Compared to the Softmax baseline (chart 1), the unification scheme (chart 2) generates larger objective values. Due to a better fit to the data distribution, validation error follows training error, quite closely and outperforms the baseline of Softmax. A higher level of validation energy (chart3) generates better performance, when the unified backpropagation is employed. With respect to MNIST dataset, some high fluctuation spots appear in the validation errors, but the overall trend shows a reasonably smooth decay rate.}
\label{cifar}
\end{figure}

Figures~\ref{mnist} and~\ref{cifar} present the comparative plots for MNIST and CIFAR-10, respectively. It is obvious that the unified backpropagation provides better generalization, and improves the training-test errors of the classification task. In Figure~\ref{mnist}, the gap between training and validation errors, hugely reduces by the proposed unification method. It means that, this provides better generalization for the trained model. the energy of loss function for Softmax is lower than the proposed method. Although this might result in the better performance for the former, the latter outperforms, in regards, to the test errors. The reason lies in the capability of this method to deal with non-smooth decision boundaries of non-convex objectives in deep neural networks. This is a critical point, especially when the classes are highly correlated in the datasets, as they are for MNSIT or CIFAR-10.

\begin{table}[t]
\begin{center}
\begin{tabular}{|c||c|c||c|c|}
\hline
\multicolumn{1}{|c||}{\multirow{2}{*}{Dataset}} & \multicolumn{2}{c||}{Softmax} & \multicolumn{2}{c|}{Unified}\\
\cline{2-5}
& \multicolumn{1}{c|}{Train (\%)} & \multicolumn{1}{c||}{Test (\%)} & \multicolumn{1}{c|}{Train (\%)} & \multicolumn{1}{c|}{Test (\%)} \\
\hline\hline
MNIST & 0.09 & 0.65 & \textbf{0.04} & \textbf{0.32}\\
CIFAR-10 & \textbf{1.32} & 22.72 & 1.56 & \textbf{18.77}\\
CIFAR-100 & \textbf{0.17} & 50.90 & 0.21 & \textbf{48.01}\\
SVHN & 0.13 & 3.81 & \textbf{0.07} & \textbf{2.59}\\
\hline
\end{tabular}
\end{center}
\caption[Classification errors for Softmax and the unified backpropagation.]{Classification errors for Softmax and the unified backpropagation. Bold values indicate the minimum training-test errors for each datasets. The proposed unification approach outperforms in test errors. Softmax produces greater training precisions for CIFAR datasets, than MNIST and SVHN.}
\label{tab:single-loss-Softmax}
\vfill
\begin{center}
\begin{tabular}{|c||c|c||c|c|}
\hline
\multicolumn{1}{|c||}{\multirow{2}{*}{Dataset}} & \multicolumn{2}{c||}{SVM} & \multicolumn{2}{c|}{Unified}\\
\cline{2-5}
& \multicolumn{1}{c|}{Train (\%)} & \multicolumn{1}{c||}{Test (\%)} & \multicolumn{1}{c|}{Train (\%)} & \multicolumn{1}{c|}{Test (\%)} \\
\hline\hline
MNIST & \textbf{0.07} & 0.53 & 0.08 & \textbf{0.51}\\
CIFAR-10 & \textbf{1.27} & 19.87 & 1.47 & \textbf{17.64}\\
CIFAR-100 & \textbf{0.13} & 49.11 & 0.17 & \textbf{39.76}\\
SVHN & 0.12 & 3.46 & \textbf{0.09} & \textbf{2.38}\\
\hline
\end{tabular}
\end{center}
\caption[Classification errors for SVM and the unified backpropagation.]{Classification errors for SVM and the unified backpropagation. Bold values indicate the minimum training-test errors for each datasets. The unification paradigm gives the best test performances, but SVM outperforms on training errors for all datasets, except SVHN.}
\label{tab:single-loss-svm}
\vfill
\begin{center}
\begin{tabular}{|c||c|c||c|c|}
\hline
\multicolumn{1}{|c||}{\multirow{2}{*}{Dataset}} & \multicolumn{2}{c||}{LDA} & \multicolumn{2}{c|}{Unified}\\
\cline{2-5}
& \multicolumn{1}{c|}{Train (\%)} & \multicolumn{1}{c||}{Test (\%)} & \multicolumn{1}{c|}{Train (\%)} & \multicolumn{1}{c|}{Test (\%)} \\
\hline\hline
MNIST & 0.07 & 0.42 & \textbf{0.05} & \textbf{0.38}\\
CIFAR-10 & 0.94 & 7.39 & \textbf{0.87} & \textbf{6.95}\\
CIFAR-100 & 0.15 & 19.63 & \textbf{0.13} & \textbf{18.46}\\
SVHN & 0.06 & 2.78 & \textbf{0.06} & \textbf{2.27}\\
\hline
\end{tabular}
\end{center}
\caption[Classification errors for LDA and the unified backpropagation.]{Classification errors for LDA and the unified backpropagation. Bold values indicate the minimum training-test errors for each datasets. The best performances come from the proposed unification scheme.}
\label{tab:single-loss-lda}
\end{table}

Figure~\ref{cifar} shows that the generalization for CIFAR-10, imposed by the unified backpropagation, is much higher than MNIST. As mentioned before, less correlation between classes in CIFAR-100 dataset, results in a better job at tuning the model parameters by this backpropagation strategy. Another observation for CIFAR-10 is that, the pattern of variations in the baseline and outcomes, is not highly aligned with MNIST. It seems that in some epochs, the unified backpropagation forces the learning process towards special classes, which do not contribute to the overall precision of classification.

\subsection{Multi-Objective Learning}
\label{sec:multiobjective}

This scenario applies the unified backpropagation to combine Softmax, SVM, and LDA objective functions. The baselines are produced by ensemble learning via Adaboost algorithm. Tables~\ref{tab:hybrid-loss-Softmax-svm},~\ref{tab:hybrid-loss-Softmax-lda}, and~\ref{tab:hybrid-loss-Softmax-svm-lda} summarize the training-test errors on all the experimental datasets, for Softmax + SVM, Softmax + LDA, and Softmax + SVM + LDA, powered by the proposed backpropagation paradigm. It can be seen that, almost everywhere, the proposed unification improves the classification performance. The only exception is Softmax + LDA on CIFAR-100 dataset, where this method is not able to outperform the baseline.

Table~\ref{tab:hybrid-loss-Softmax-svm} shows the outcomes of the unification on Softmax and SVM combination. It is obvious that, the unified backpropagation improves test errors for all the experiments. The training errors are all improved, except on CIFAR-100 dataset. In table~\ref{tab:hybrid-loss-Softmax-lda}, we report the errors for the joint Softmax and LDA. Here, all the improvements come from the unified backpropagation. On CIFAR-100, training error increases from 0.9\% to 1.5\%, and on CIFAR-100 the testing error jumps from 18.27\% to 35.28\%. Since these degradations belong to CIFAR, it can be concluded that, LDA is not that successful at separating their highly-correlated classes.

Finally, Table~\ref{tab:hybrid-loss-Softmax-svm-lda} gathers the results of experiments for the composition of Softmax, SVM, and LDA. It is clear that the unification scheme outperforms the baseline on all training-test errors, except for the training on CIFAR-10 and test on CIFAR-100 datasets. It seems that SVM makes a significant contribution towards compensating LDA's disadvantage on CIFAR datasets, but that is not enough for the unified backpropagation to take an edge over baseline.  

\begin{table}[t]
\begin{center}
\begin{tabular}{|c||c|c||c|c|}
\hline
\multicolumn{1}{|c||}{\multirow{2}{*}{Dataset}} & \multicolumn{2}{c||}{Softmax + SVM} & \multicolumn{2}{c|}{Unified}\\
\cline{2-5}
& \multicolumn{1}{c|}{Train (\%)} & \multicolumn{1}{c||}{Test (\%)} & \multicolumn{1}{c|}{Train (\%)} & \multicolumn{1}{c|}{Test (\%)} \\
\hline\hline
MNIST & 0.08 & 0.57 & \textbf{0.07} & \textbf{0.45}\\
CIFAR-10 & 1.54 & 18.72 & \textbf{1.15} & \textbf{15.82}\\
CIFAR-100 & \textbf{0.23} & 48.85 & 0.91 & \textbf{38.58}\\
SVHN & 0.11 & 3.27 & \textbf{0.08} & \textbf{2.48}\\
\hline
\end{tabular}
\end{center}
\caption[Classification errors for Softmax + SVM and the unified backpropagation.]{Classification errors for Softmax + SVM and the unified backpropagation. Bold values indicate the minimum training-test errors for each datasets. The test errors show considerable improvements over the baseline performances.}
\label{tab:hybrid-loss-Softmax-svm}
\vfill
\begin{center}
\begin{tabular}{|c||c|c||c|c|}
\hline
\multicolumn{1}{|c||}{\multirow{2}{*}{Dataset}} & \multicolumn{2}{c||}{Softmax + LDA} & \multicolumn{2}{c|}{Unified}\\
\cline{2-5}
& \multicolumn{1}{c|}{Train (\%)} & \multicolumn{1}{c||}{Test (\%)} & \multicolumn{1}{c|}{Train (\%)} & \multicolumn{1}{c|}{Test (\%)} \\
\hline\hline
MNIST & 0.07 & 0.44 & \textbf{0.05} & \textbf{0.41}\\
CIFAR-10 & \textbf{0.90} & 7.51 & 1.50 & \textbf{6.81}\\
CIFAR-100 & \textbf{0.18} & \textbf{18.27} & 0.71 & 35.28\\
SVHN & 0.09 & 3.64 & \textbf{0.06} & \textbf{2.41}\\
\hline
\end{tabular}
\end{center}
\caption[Classification errors for Softmax + LDA and the unified backpropagation.]{Classification errors for Softmax + LDA and the unified backpropagation. Bold values indicate the minimum training-test errors for each datasets. This approach improves the precision, when compared with the baseline for all datasets, except CIFAR-100.}
\label{tab:hybrid-loss-Softmax-lda}
\vfill
\begin{center}
\begin{tabular}{|c||c|c||c|c|}
\hline
\multicolumn{1}{|c||}{\multirow{2}{*}{Dataset}} & \multicolumn{2}{c||}{Softmax + SVM + LDA} & \multicolumn{2}{c|}{Unified}\\
\cline{2-5}
& \multicolumn{1}{c|}{Train (\%)} & \multicolumn{1}{c||}{Test (\%)} & \multicolumn{1}{c|}{Train (\%)} & \multicolumn{1}{c|}{Test (\%)} \\
\hline\hline
MNIST & 0.08 & 0.38 & \textbf{0.05} & \textbf{0.30}\\
CIFAR-10 & \textbf{0.78} & 5.96 & 1.83 & \textbf{5.44}\\
CIFAR-100 & 1.35 & \textbf{22.49} & \textbf{0.68} & 35.13\\
SVHN & 0.08 & 3.01 & \textbf{0.07} & \textbf{2.34}\\
\hline
\end{tabular}
\end{center}
\caption[Classification errors for Softmax + SVM + LDA and the unified backpropagation.]{Classification errors for Softmax + SVM + LDA and the unified backpropagation. Bold values indicate the minimum training-test errors for each datasets. This paradigm outperforms the baseline on almost all the experimental datasets, except CIFAR-100.}
\label{tab:hybrid-loss-Softmax-svm-lda}
\end{table}

All in all, LDA does a better job than SVM in both of the baseline and unified backpropagation, and the proposed method performs better, when all the objectives come together. The best improvement goes to CIFAR-100 dataset, which reduces the test error from 38.58\% for Softmax + SVM to 35.28\% for Softmax + LDA, followed by 35.13\% for Softmax + SVM + LDA. For CIFAR-10, Softmax + LDA improves the performance quite well, in comparison with, Softmax + SVM. Although the joint venture of all classifiers, generates higher precisions in test, the lowest training errors, varies between CIFAR-10 for Softmax + SVM, MNIST and SVHN for Softmax + LDA, and CIFAR-100 for Softmax + SVM + LDA.

\newpage

\subsection{Discussion}
\label{sec:discussion}

\begin{table}[!t]
\begin{center}
\begin{tabular}{|c||c|c||c|c|}
\hline
\multicolumn{1}{|c||}{\multirow{2}{*}{Dataset}} & \multicolumn{2}{c||}{Baseline} & \multicolumn{2}{c|}{Unified}\\
\cline{2-5}
& \multicolumn{1}{c|}{Train (\%)} & \multicolumn{1}{c||}{Test (\%)} & \multicolumn{1}{c|}{Train (\%)} & \multicolumn{1}{c|}{Test (\%)} \\
\hline\hline
MNIST & 0.08 & 0.38 & \textbf{0.05} & \textbf{0.30}\\
CIFAR-10 & \textbf{0.94} & 7.39 & 1.83 & \textbf{5.44}\\
CIFAR-100 & 0.18 & \textbf{18.27} & \textbf{0.13} & 18.46\\
SVHN & 0.07 & 2.78 & \textbf{0.06} & \textbf{2.27}\\
\hline
\end{tabular}
\end{center}
\caption[Minimum test errors for the baseline and unified backpropagation.]{Minimum test errors for the baseline and  unified backpropagation. Bold values indicate the minimum training-test errors for each datasets. The unification strategy is consistently successful at improving the classification performance of deep convolutional networks.}
\label{tab:benchmark}
\end{table}

Considering both of the experimental scenarios, Table~\ref{tab:benchmark} summarizes the minimum test errors, and its corresponding training errors for each of the datasets under examination. The unified backpropagation either outperforms baselines by high margins (5.44\% vs 7.39\% for CIFAR-10) or follows them by close rates (18.46\% vs 18.27\% for CIFAR-100). This confirms  advantage of the proposed backpropagation for the classification.

On the other hand, the best results for the unification method, go to multi-objective learning on MNIST \& CIFAR-100 and single-objective learning on CIFAR-100 \& SVHN datasets. Although further investigations remain, initial results indicate that multi-objective learning performs better on a small or medium number of samples-classes, while single-objective learning performs best on a large number of samples-classes. This is due to the fact that the multi-objective regime is not able to cope, with either complex data distributions, or highly-correlated classes, when several objectives contradict each other. 

\section{Conclusion}
\label{sec:conclusion}

The typical classification architectures in deep neural networks employ Softmax, support vector machines or linear discriminant analysis as the top layer and backpropagate the error by the gradient of their specific lost functions. We propose a novel paradigm to learn hybrid multi-objective networks with unified backpropagation. Using basic probability assignment form evidence theory, we link the gradients of hybrid loss functions and update the network parameters by backpropagation. This also avoid biases in imbalanced data distributions and improves the classification performance of single-objective or hybrid models. Our extensive experiments on standard datasets prove that the proposed unification scheme contributes to the overall precision of deep convolutional neural networks.

\newpage

\bibliographystyle{nips}
\bibliography{nips}

\end{document}